\documentclass[10pt,twocolumn,letterpaper]{article}

\usepackage[pagenumbers]{wacv} 

\usepackage{graphicx}
\usepackage{amsmath}
\usepackage{amssymb}
\usepackage{booktabs}
\usepackage{color}
\usepackage{xcolor}
\usepackage{array}
\usepackage{ragged2e}
\usepackage{booktabs}
\usepackage{lipsum}
\usepackage{rotating}
\usepackage{tabularray}
\usepackage{colortbl}
\usepackage{arydshln}
\usepackage{makecell}
\usepackage{vcell}
\usepackage{multirow}
\usepackage{cellspace}
\usepackage[normalem]{ulem} 

\usepackage{algorithm}
\usepackage{algpseudocode}

\usepackage{adjustbox}

\usepackage{bm}
\usepackage{calrsfs}
\usepackage{amssymb}
\usepackage{amsmath}

\usepackage{graphicx} 
\usepackage{float}  

\usepackage[pagebackref,breaklinks,colorlinks]{hyperref}

\usepackage[capitalize]{cleveref}
\crefname{section}{Sec.}{Secs.}
\Crefname{section}{Section}{Sections}
\Crefname{table}{Table}{Tables}
\crefname{table}{Tab.}{Tabs.}

\def\wacvPaperID{1278} 
\def\confName{WACV}
\def\confYear{2025}

\begin{document}

\title{Stratified Domain Adaptation: A Progressive Self-Training Approach for\\Scene Text Recognition}

\author{Kha Nhat Le, Hoang-Tuan Nguyen, Hung Tien Tran, Thanh Duc Ngo\thanks{Corresponding author}\\
University of Information Technology, VNU-HCM, Vietnam\\
Vietnam National University, Ho Chi Minh City, Vietnam\\
{\tt\small \{20520208, 20520344, 19521587\}@gm.uit.edu.vn, thanhnd@uit.edu.vn}
}

\maketitle

\begin{abstract}
Unsupervised domain adaptation (UDA) has become increasingly prevalent in scene text recognition (STR), especially where training and testing data reside in different domains. The efficacy of existing UDA approaches tends to degrade when there is a large gap between the source and target domains. To deal with this problem, gradually shifting or progressively learning to shift from domain to domain is the key issue. In this paper, we introduce the Stratified Domain Adaptation (StrDA) approach, which examines the gradual escalation of the domain gap for the learning process. The objective is to partition the training data into subsets so that the progressively self-trained model can adapt to gradual changes. We stratify the training data by evaluating the proximity of each data sample to both the source and target domains. We propose a novel method for employing domain discriminators to estimate the out-of-distribution and domain discriminative levels of data samples. Extensive experiments on benchmark scene-text datasets show that our approach significantly improves the performance of baseline (source-trained) STR models.
\end{abstract}

\section{Introduction}
\label{sec:intro}
Although recent STR models have shown impressive performance, they are typically trained exclusively on labeled synthetic data. Baek \etal \cite{baek2021if} have emphasized the significance of training STR models on real data, asserting its greater importance compared to synthetic data. However, collecting labeled real data poses a considerable challenge because of its high cost and time-intensive nature. Some efforts have been dedicated to generating synthetic data that closely resembles real data. The problem remains challenging due to the domain gap. Significant performance degradation occurs when a model trained with synthetic data is applied to real data due to the substantial disparity between data distributions across domains. To address this problem, domain adaptation approaches are proposed to reduce distribution offsets. Especially, learning approaches that involve gradually shifting or progressively learning have demonstrated more notable improvement than learning directly from one source domain into another target domain.

In addition, while labeled real data is scarce, unlabeled real data is abundant and easily collectible. Several approaches used self-training methods to harness both labeled synthetic and unlabeled real data, with the aim of improving model performance \cite{janouskova2021text, baek2021if, patel2021feds, li2021domain, sun2022semi, li2023fine, song2023mugs}. This has demonstrated considerable effectiveness in improving the performance of the model and enabling the model to leverage the latent knowledge from unlabeled data points. However, it comes with several drawbacks due to its inherent instability. There is no explicit guarantee of the accuracy of the pseudo-labeling, which could cause the model to degrade. As the classification model weakens or the gap between the source and target domains increases, this phenomenon becomes more pronounced, making it challenging to control the upper bound on self-training errors \cite{kumar2020understanding}.

In this work, we propose the Stratified Domain Adaptation (StrDA) approach by leveraging the gradual escalation of the domain gap to effectively address the discrepancy between the source and target domains. We partition the learning set from the target domain into smaller subsets, such that the domain gap of each subset compared to the source domain progressively increases. This way, the model can gradually adapt to domain changes and improve its performance. To evaluate the proximity of each data sample to the source and target domains, we propose the Harmonic Domain Gap Estimator ($\mathrm{HDGE}$), which employs a pair of discriminators. Each discriminator evaluates the out-of-distribution (OOD) levels for each data point, with particular reference to the source or target domain. Then, these two OOD-level evaluations are passed through a harmonic function to estimate the distance from the data to the source domain. This means that data points that are situated near the intersection of the two domains exhibit a minimal distance, while data points that are outside the distribution of both domains display an exceedingly large distance. The harmonic evaluation function provides a more precise assessment of out-of-distribution levels, ensuring a more reliable assessment of OOD data points.

We summarize our contributions as follows:

 \begin{itemize}
     \item We introduce a progressive self-training domain adaptation approach for scene text recognition, which helps improve the model's performance by utilizing unlabeled data with high-quality pseudo-labels. We propose the Harmonic Domain Gap Estimator method for stratifying domain gaps by analyzing the gradual escalation of domain gaps, which plays a crucial role in the effectiveness of domain adaptation.

     \item Extensive experiments are conducted on six benchmark datasets (IIIT \cite{mishra2012scene}, SVT\cite{wang2011end}, IC13 \cite{karatzas2013icdar}, IC15 \cite{karatzas2015icdar}, SVTP \cite{phan2013recognizing}, CUTE \cite{risnumawan2014robust}) and five additional datasets COCO \cite{veit2016coco}, Uber \cite{zhang2017uber}, ArT \cite{chng2019icdar2019}, ReCTS \cite{zhang2019icdar}, Union14M \cite{jiang2023revisiting}) to assess the performance of the proposed approach. It leads to a significant improvement in the performance of various existing STR models. This paves the way for recognizing text without incurring human annotation costs, particularly in cases where labeled real data is limited.
\end{itemize}

\begin{figure*}[tb]
    \centering
    \includegraphics[width=\textwidth]{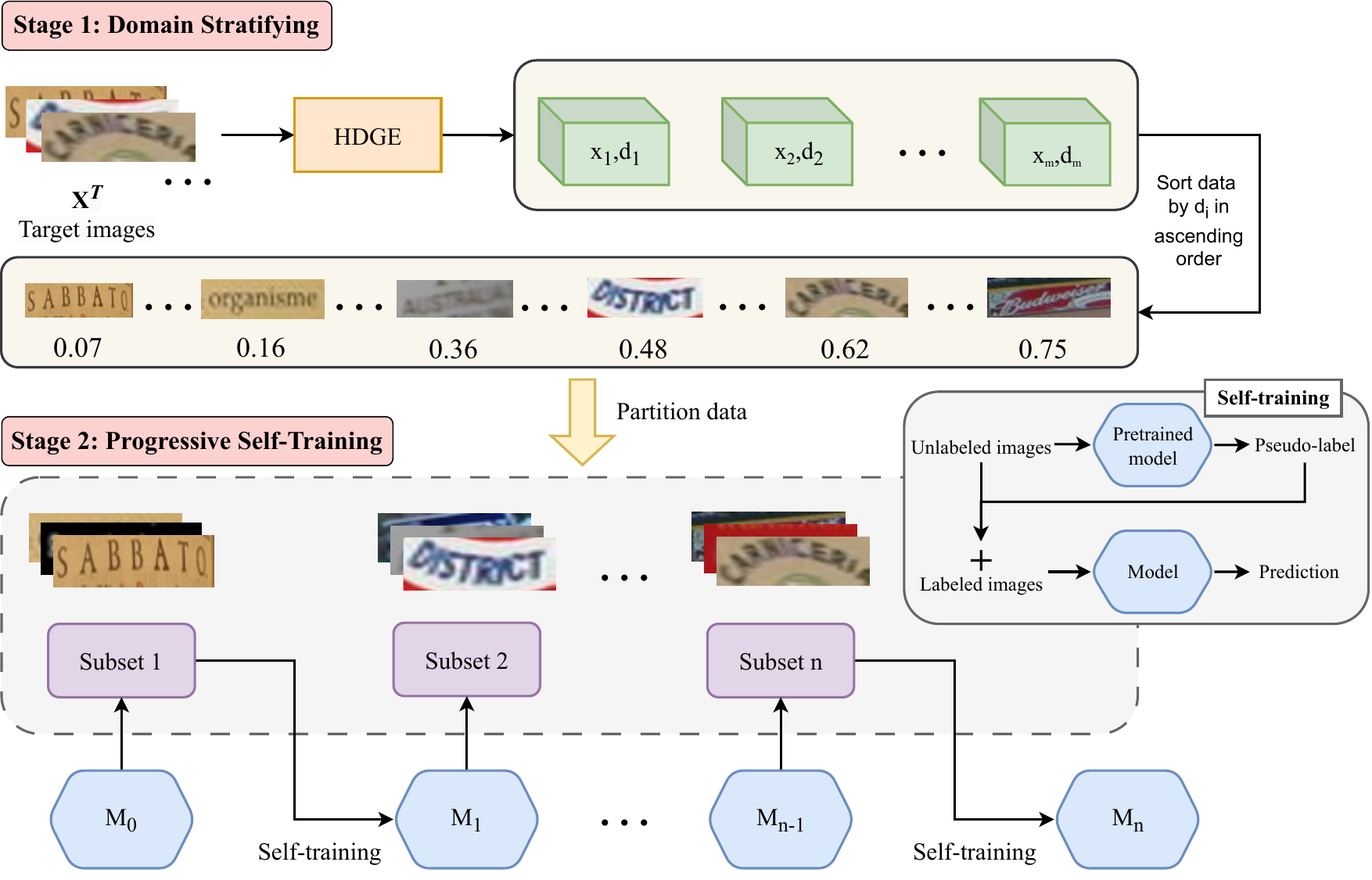}
    \caption{\textbf{The overall framework of our proposed Stratified Domain Adaptation (StrDA) for scene text recognition.} Our approach leverages labeled synthetic data and unlabeled real data, without human annotation. The entire process is divided into 2 stages: Domain Stratifying (partitioning the unlabeled real data into subsets satisfying \cref{eq:domain}) and Progressive Self-Training. $m$ represents the number of unlabeled data, and $n$ serves as the hyper-parameter in \cref{eq:domain}.}
    \label{fig:Overview}
\end{figure*}

\section{Related Work}
\label{sec:related}
\subsection{Scene Text Recognition}

In general, Scene Text Recognition (STR) is treated as a sequence prediction task that utilizes sequence modeling to leverage robust visual features for recognition. The CTC-based \cite{graves2006connectionist} decoder methods \cite{shi2016end, he2016reading, atienza2021vision, du2022svtr} aim to maximize the probability of all possible paths for the final prediction, achieving a balance between accuracy and efficiency. Attention-based decoder methods \cite{lee2016recursive, shi2016robust, cheng2017focusing, shi2018aster, yue2020robustscanner, lee2020recognizing, wang2020decoupled, wang2022multi, zhang2023linguistic, cheng2023lister} utilize a visual query to localize the position of each character via an attention mechanism with the idea inspired from NLP Community \cite{cheng2016long, vaswani2017attention}. This approach has demonstrated robustness in precision, albeit with high computational costs. Furthermore, some studies \cite{fang2021read, na2022multi, da2022levenshtein, zhao2023clip4str} have shown the effectiveness of integrating the external language model to capture text semantics. 

Most of the STR methods mentioned above have achieved remarkable results on common benchmarks, even when trained solely on synthetic datasets in a supervised manner. However, there is a significant domain gap between synthetic and real-world data. Jiang \etal \cite{jiang2023revisiting} identified that STR is far from being solved by analyzing the numerous challenges associated with real-world data from a data-oriented perspective.

\subsection{Domain Adaptation for Scene Text Recognition}

Domain Adaptation (DA) is a technique designed to enhance the performance of models trained on the source domain when applied to the target domain. One widely used approach in DA is self-training \cite{lee2013pseudo}. The self-training process involves training a model with labeled data, then using this model to generate pseudo-labels for unlabeled data, which are subsequently used to retrain the target model. However, pseudo-labeling (PL) is often suboptimal due to erroneous predictions from poorly calibrated models, which can negatively impact training efficiency \cite{pham2021metapseudolabels, arazo2020pseudo}. Recent works \cite{xie2020self, pham2021metapseudolabels, rizve2021defense} have focused on reducing PL errors in self-learning for general tasks and have demonstrated its effectiveness in more specific applications as well.

Recently, various studies have focused on refining the PL processes to harness labeled synthetic data and unlabeled real data in STR. Baek \etal \cite{baek2021if} explored multiple ways to enhance STR models by using pseudo-labels. Patel \etal \cite{patel2023seq} introduced an uncertainty-based label selection strategy for STR by utilizing Beam-Search inference. Fang \etal \cite{fang2021read} proposed the Ensemble Self-training strategy by treating the iterative predictions as an ensemble. Li \etal \cite{li2023fine} introduced the Adaptive Distribution Regularizer to bridge the domain gap and achieved remarkable performance in cross-domain adaptation with both scene text and handwritten text. In another alternative perspective, several works \cite{zhan2019ga, zhang2019sequence, zhang2021robust, chang2022smile, tien2023unsupervised, liu2023unsupervised, liu2023protouda} have recently suggested domain adaptation techniques to learn feature discrepancy between source and target domains. Zhang \etal \cite{zhang2021robust} employs an Adversarial Sequence-to-Sequence Domain Adaptation (ASSDA) network, which could adaptively align the coarse global-level and fine-grained character-level representation across domains in an adversarial manner. Liu \etal \cite{liu2023protouda} introduced ProtoUDA, which enhances text recognition across various domains by using pseudo-labeled character features and parallel, complementary modules for class-level and instance-level alignment. 

The inherent domain gap between labeled and unlabeled data mainly results in low-quality derived pseudo-labels. Although these methods offer promising results, they primarily focus on addressing the domain gap through direct adaptation. We take into account that the domain gap has a progressive tendency. Instead of directly adapting from the source to the target domain, we propose to leverage the gradual shift between domains in scene text recognition. Recent studies in other computer vision tasks have also presented approaches such as Gradual Adaptation \cite{dai2018dark, gadermayr2018gradual, chen2021gradual, huang2023divide} and Easy-to-Hard Transfer \cite{chen2019progressive, westfechtel2024gradual} for adapting to different domains. These approaches demonstrate significant performance enhancement compared to direct UDA.

\section{Stratified Domain Adaptation}
\label{sec:method}
\subsection{Overview}
\label{sec:method}

\begin{figure*}[tb]
    \centering
    \includegraphics[width=0.84\linewidth]{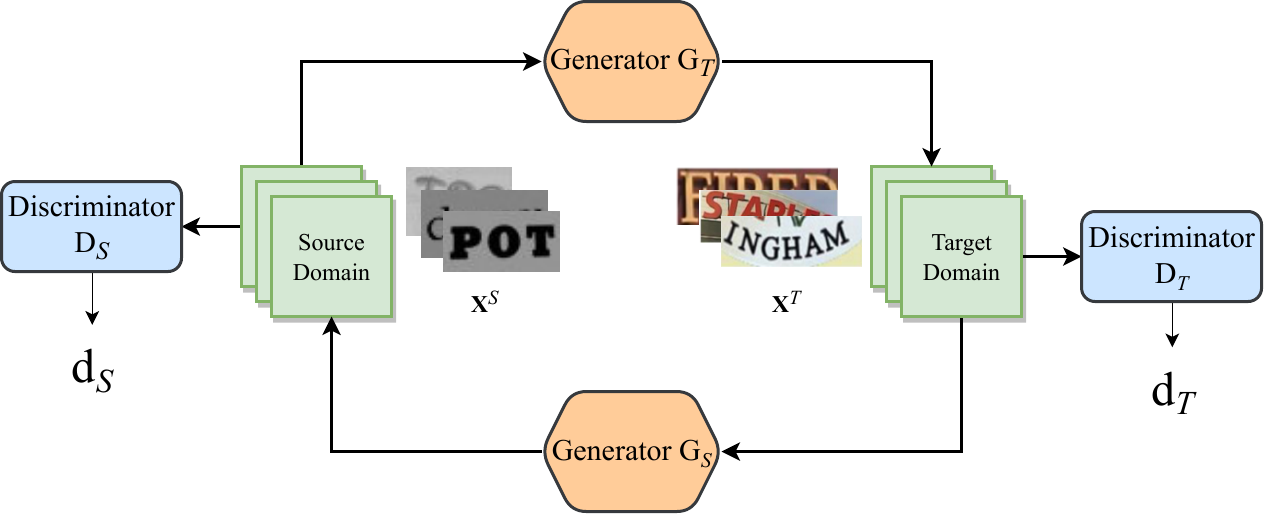}
    \caption{Our architecture consists of two mapping functions, $G_T: S \rightarrow T$ and $G_S: T \rightarrow S$, along with associated adversarial discriminators, $D_S$ and $D_T$. While $G_S$ and $G_T$ are tasked with translating images from one domain to another, $D_S$ estimates the difference between an image and the data distribution of the source domain $S$, and similarly, $D_T$ does so for the target domain $T$.}
    \label{fig:cyclegan}
\end{figure*}

In this work, our focus is on addressing the problem using two predefined datasets: one comprising labeled data samples from the source domain, $S=\left\{(\bm{x}_i^{S} \bm{y}_i^{S})\right\}_{i=1}^{|S|}$; and, the other comprising unlabeled data samples from the target domain, $T=\left\{\bm{x}_i^{T}\right\}_{i=1}^{|T|}$. The goal of domain adaptation is to enhance the performance of the source-trained model by leveraging both $S$ and $T$.

\textbf{Unsupervised Domain Adaptation (UDA)}. To investigate the Stratified Domain Gap approach, we relied on the traditional UDA approach using \textit{vanilla self-training} (denoted as $\mathrm{ST}$). $\mathrm{ST}$ takes a source-trained model (referred to as the \textit{baseline model}) to generate a pseudo-label for $\bm{x}_i^{T}$. Subsequently, the model is trained using the pseudo-labeled data combined with the labeled data from the source domain. Applying domain adaptation directly (using the entire dataset for a single self-training process) may encounter several disadvantages (\cref{sec:intro}). Instead, our approach employs a series of $\mathrm{ST}$ rounds with a sequence of target sub-domain data.

We first partition the unlabeled data into a sequence of equally-sized subsets $T_{1}, T_{2}, T_{3},\ldots, T_{n}$, where $T_{m}=\left\{{\bm{x}}_i^{T_{m}}\right\}_{i=1}^{|T_{m}|}$. By this, we assume that the domain gap between $T_{m}$ and $S$ is less than that between $T_{m+1}$ and $S$:
\begin{align}
    \rho (S, T_{m}) \leq \rho (S, T_{m+1}), \quad \forall m \in (1,n)
    \label{eq:domain}
\end{align}
where $\rho (P, Q)$\footnote{$p$ can be Kullback-Leibler Divergence (KL Divergence) or Wasserstein Distance} is a distance function between distributions $P$ and $Q$.

To partition the data with respect to \cref{eq:domain}, we propose the Harmonic Domain Gap Estimator ($\mathrm{HDGE}$) method to estimate the proximity of a data point $\bm{x_i} \in T$ and the source domain $S$. Afterward, we arrange and partition the data that satisfy \cref{eq:domain}. We refer to the entire process as Stage 1-Domain Stratifying. After obtaining the subsets from stage 1, we sequentially apply $\mathrm{ST}$ to each subset. This process is referred to as Stage 2-Progressive Self-Training. The entire Stratified Domain Adaptation approach consists of two stages, as described in \cref{fig:Overview}.

\subsection{Stage 1: Domain Stratifying}
\label{method:stage1}

Given $S$ and $T$, we introduce $\mathrm{HDGE}$ to assess the proximity of a data point $\bm{x}_{i}^T \in T$ and $S$, denoted as $d_{i}$. Lower $d_{i}$ indicates that $\bm{x}_{i}^T$ is closer to the source. After assigning $d_{i}$ to each data point $\bm{x}_{i}^T$, we arrange the data in ascending order of $d_{i}$ and then partition them into $n$ subsets with equal size, $T_{m}=\left\{{\bm{x}}_i^{T_{m}}\right\}_{i=1}^{|T_{m}|}$, for progressive self-training.

\textbf{Harmonic Domain Gap Estimator ($\mathrm{HDGE}$)} uses a pair of discriminators, one for the source domain and the other for the target domain ($D_S$ and $D_T$). Each discriminator evaluates the out-of-distribution (OOD) levels for each data point. By synthesizing the outputs of the two discriminators, we can determine whether the data is in-domain (near source or target) or out of both distributions. We denote these two OOD levels as $d_S$ and $d_T$. To calculate the $d_{i}$ for $\bm{x}_i^{T}$, we use the formula:

\begin{equation}
    \begin{aligned}
    d_i = \frac{(1+\beta^2).d_S(\bm{x}_i^{T}).d_T(\bm{x}_i^{T})}{\beta^2.d_S(\bm{x}_i^{T})+d_T(\bm{x}_i^{T})}
    \end{aligned}
    \label{eq:hdge}
\end{equation}
where $0 \le \beta<1$, we tend to bias the data towards smaller $d_{S}(\bm{x}_i^{T})$, meaning closer to the source domain. This aligns with the condition \cref{eq:domain}.

With the designed $d_i$-computation function as above, we aim to arrange the data for the progressive self-training process with the following prioritization:

\begin{enumerate}
    \item $\bm{x_i}$ situated at intersection of two distributions ($d_S$ and $d_T$ are small)
    \item $\bm{x_i}$ closer to the source domain (small $d_S$, large $d_T$)
    \item $\bm{x_i}$ closer to the target domain (small $d_T$, large $d_S$)
    \item $\bm{x_i}$ that is out of two distributions ($d_S$ and $d_T$ are large)
\end{enumerate}

To create a pair of discriminators $D_S$ and $D_T$ with the ability to assess out-of-distribution (OOD) levels effectively, we designed a learning strategy inspired by \cite{zhu2017unpaired}. As illustrated in \cref{fig:cyclegan}, in addition to the two discriminators $D_S$ and $D_T$, we also utilize two generators: $G_T$ translates images from the source domain to the target domain ($G_{T}: S \rightarrow T$), and $G_S$ performs a similar task from the target domain to the source domain ($G_{S}: T \rightarrow S$).

While \textit{generators} strive to learn how to represent from one domain to another, \textit{discriminators} learn to distinguish between images generated by the generator and \textit{real} images. Through adversarial learning, $G_S$ and $G_T$ will improve image generation, consequently enhancing the discriminative abilities of $D_S$ and $D_T$. As a result, when a new data point $x_i$ is introduced, the discriminator pair accurately assesses out-of-distribution levels ($d_S$ and $d_T$).

Given training samples \(\left\{\bm{x}_i^{S}\right\}_{i=1}^{|S|}\) where \(\bm{x}_i^{S} \in S\) and \(\left\{\bm{x}_i^{T}\right\}_{i=1}^{|T|}\) where \(\bm{x}_i^{T} \in T\), the data distribution is indicated as \(\bm{x}^{S} \sim p_{\text{data}}(\bm{x}^{S})\) and \(\bm{x}^{T} \sim p_{\text{data}}(\bm{x}^{T})\). The adversarial loss for the mapping function $G_T: S \rightarrow T$ and its discriminator $D_T$ is expressed as follows:
\begin{equation}
    \begin{split}
    L_{GAN}(G_T, D_T, S, T) = \mathbb{E}_{\bm{x}^{T} \sim p_{\text {data}}(\bm{x}^{T})}[\log D_T(\bm{x}^{T})] \\ +\mathbb{E}_{\bm{x}^{S} \sim p_{\text {data }}(\bm{x}^{S})}[\log(1-D_T(G_T(\bm{x}^{S})))]
    \end{split}
\end{equation}
where $G_T$ attempts to generate images $G_T(\bm{x}^{S})$ that resemble images from domain $T$, while the objective of $D_T$ is to differentiate between translated samples $G_T(\bm{x}^{S})$ and real samples $\bm{x}^{T}$. $G_T$ strives to minimize this objective against the adversary $D_T$ that seeks to maximize it, \ie $min_{G_T} max_{D_T} L_{GAN}(G_T, D_T, S, T)$. We use a similar adversarial loss for the mapping function $G_S: T \rightarrow S$ and its discriminator $D_S$ as well: \ie $min_{G_S} max_{D_S} L_{GAN}(G_S, D_S, T, S)$

After training, we obtain a pair of discriminators, $D_S$ and $D_T$ with the ability to estimate the domain gap $d_i$ for data $\bm{x}_i^{T}$ using \cref{eq:hdge}.

\definecolor{Gallery}{rgb}{0.941,0.941,0.941}
\definecolor{Mercury}{rgb}{0.89,0.89,0.89}
\definecolor{Green}{rgb}{0,0.713,0}
\definecolor{JapaneseLaurel}{rgb}{0,0.443,0}
\begin{table*}[t]
\caption{\textbf{Word accuracy on six scene-text benchmarks and five additional datasets}. The number of words in each dataset is listed with its name. "(baseline)" means models trained only on synthetic data. We present the results of domain adaptation approaches for the baseline model. "$\mathrm{ST}$" refers to traditional unsupervised domain adaptation (vanilla self-training). "$\mathrm{StrDA_{HDGE}}$" is Stratified Domain Adaptation methods using Harmonic Domain Gap Estimator. The numbers denoted by "$\mathrm{\Delta}$" in \textcolor[rgb]{0,0.718,0}{\textbf{green}} indicate improvements over each dataset. With each dataset (\ie each column), the best result is shown in \textbf{bold}. With each baseline model on a dataset, the best result is shown with an \uline{underline}. The performance of baseline models is substantially enhanced by the proposed methods.}
\centering
\adjustbox{max width=\textwidth}
{
\newcommand{\plus}{\raisebox{.25\height}{\scalebox{.8}{\textbf{+}}}}
\begin{tblr}{
  row{1} = {c},
  column{1} = {l},
  column{2} = {l},
  column{4} = {c},
  column{5} = {c},
  column{6} = {c},
  column{7} = {c},
  column{8} = {c},
  column{9} = {c},
  column{11} = {c},
  column{12} = {c},
  column{13} = {c},
  column{14} = {c},
  cell{1}{1} = {r=2}{b},
  cell{1}{2} = {r=2}{b},
  cell{1}{3} = {c=7}{},
  cell{1}{10} = {c=5}{},
  cell{2}{3} = {c},
  cell{2}{9} = {r=2}{},
  cell{2}{10} = {c},
  cell{3}{3} = {c},
  cell{3}{10} = {c},
  cell{4}{3} = {c},
  cell{4}{10} = {c},
  cell{5}{3} = {c},
  cell{5}{10} = {c},
  cell{6}{2} = {c},
  cell{6}{3} = {c,fg=Green},
  cell{6}{4} = {fg=Green},
  cell{6}{5} = {fg=Green},
  cell{6}{6} = {fg=Green},
  cell{6}{7} = {fg=Green},
  cell{6}{8} = {fg=Green},
  cell{6}{9} = {fg=Green},
  cell{6}{10} = {c,fg=Green},
  cell{6}{11} = {fg=Green},
  cell{6}{12} = {fg=Green},
  cell{6}{13} = {fg=Green},
  cell{6}{14} = {fg=Green},
  cell{7}{3} = {c},
  cell{7}{10} = {c},
  cell{8}{2} = {c},
  cell{8}{3} = {c,fg=Green},
  cell{8}{4} = {fg=Green},
  cell{8}{5} = {fg=Green},
  cell{8}{6} = {fg=Green},
  cell{8}{7} = {fg=Green},
  cell{8}{8} = {fg=Green},
  cell{8}{9} = {fg=Green},
  cell{8}{10} = {c,fg=Green},
  cell{8}{11} = {fg=Green},
  cell{8}{12} = {fg=Green},
  cell{8}{13} = {fg=Green},
  cell{8}{14} = {fg=Green},
  cell{9}{3} = {c},
  cell{9}{10} = {c},
  cell{10}{3} = {c},
  cell{10}{10} = {c},
  cell{11}{2} = {c},
  cell{11}{3} = {c,fg=Green},
  cell{11}{4} = {fg=Green},
  cell{11}{5} = {fg=Green},
  cell{11}{6} = {fg=Green},
  cell{11}{7} = {fg=Green},
  cell{11}{8} = {fg=Green},
  cell{11}{9} = {fg=Green},
  cell{11}{10} = {c,fg=Green},
  cell{11}{11} = {fg=Green},
  cell{11}{12} = {fg=Green},
  cell{11}{13} = {fg=Green},
  cell{11}{14} = {fg=Green},
  cell{12}{3} = {c},
  cell{12}{10} = {c},
  cell{13}{2} = {c},
  cell{13}{3} = {c,fg=Green},
  cell{13}{4} = {fg=Green},
  cell{13}{5} = {fg=Green},
  cell{13}{6} = {fg=Green},
  cell{13}{7} = {fg=Green},
  cell{13}{8} = {fg=Green},
  cell{13}{9} = {fg=Green},
  cell{13}{10} = {c,fg=Green},
  cell{13}{11} = {fg=Green},
  cell{13}{12} = {fg=Green},
  cell{13}{13} = {fg=Green},
  cell{13}{14} = {fg=Green},
  cell{14}{3} = {c},
  cell{14}{10} = {c},
  cell{15}{3} = {c},
  cell{15}{10} = {c},
  cell{16}{2} = {c},
  cell{16}{3} = {c,fg=Green},
  cell{16}{4} = {fg=Green},
  cell{16}{5} = {fg=Green},
  cell{16}{6} = {fg=Green},
  cell{16}{7} = {fg=Green},
  cell{16}{8} = {fg=Green},
  cell{16}{9} = {fg=Green},
  cell{16}{10} = {c,fg=Green},
  cell{16}{11} = {fg=Green},
  cell{16}{12} = {fg=Green},
  cell{16}{13} = {fg=Green},
  cell{16}{14} = {fg=Green},
  cell{17}{3} = {c},
  cell{17}{10} = {c},
  cell{18}{2} = {c},
  cell{18}{3} = {c,fg=Green},
  cell{18}{4} = {fg=Green},
  cell{18}{5} = {fg=Green},
  cell{18}{6} = {fg=Green},
  cell{18}{7} = {fg=Green},
  cell{18}{8} = {fg=Green},
  cell{18}{9} = {fg=Green},
  cell{18}{10} = {c,fg=Green},
  cell{18}{11} = {fg=Green},
  cell{18}{12} = {fg=Green},
  cell{18}{13} = {fg=Green},
  cell{18}{14} = {fg=Green},
  vline{4} = {1}{},
  vline{10} = {2-18}{},
  hline{1,19} = {-}{0.08em},
  hline{2} = {3-14}{},
  hline{4,9,14} = {-}{},
  hline{5,10,15} = {2-18}{dashed},
  hline{7,12,17} = {2-18}{dashed}
}
Type      & Method                & Common Benchmarks &      &      &      &      &      &      & Additional Datasets &        &        &       &           \\
          &                       & IIIT              & SVT  & IC13 & IC15 & SVTP & CUTE & Avg. & COCO              & Uber   & ArT    & ReCTS & Union14M \\
          &                       & 3000              & 647  & 857  & 1811 & 645  & 288  &      & 9,825             & 80,418 & 35,149 & 2,592 & 403,379   \\
          & CRNN
  (baseline)     & 92.5              & 86.4 & 92.0   & 71.3 & 75.2 & 83.3 & 84.4 & 49.5              & 34.6   & 59.5   & 77.1  & 43.3      \\
          & + $\mathrm{ST}$             & \uline{93.7}              & 87.6 & 92.2 & 72.9 & 75.5 & \uline{84.7} & 85.5 & 51.4              & 35.9   & 60.7   & 79.8  & 46.2      \\
CTC       & \textbf{$\mathrm{\Delta}$} & +1.2              & +1.2 & +0.2 & +1.6 & +0.3 & +1.4 & +1.1 & +1.9              & +1.3   & +1.2   & +2.7  & +2.9      \\
          & + $\mathrm{StrDA_{HDGE}}$     & 93.4              & \uline{89.0}   & \uline{93.1} & \uline{74.0}   & \uline{77.1} & 84.4 & \uline{86.0}   & \uline{53.0}                & \uline{36.8}   & \uline{60.9}   & \uline{81.0}    & \uline{47.8}      \\
          & $\mathrm{\Delta}$ & +0.9              & +2.6 & +1.1 & +2.7 & +1.9 & +1.1 & +1.6 & +3.5              & +2.2   & +1.4   & +3.9  & +4.5      \\
          & TRBA
  (baseline)     & 96.2              & 93.7 & 95.8 & 81.9 & 86.1 & 91.0   & 91.0   & 62.5              & 39.0     & 69.0     & 82.8  & 56.6      \\
          & + $\mathrm{ST}$             & 97.1              & 94.0   & 96.1 & 82.5 & 90.1 & 92.4 & 92.0   & 65.5              & 40.9   & 70.9   & 84.8  & 60.4      \\
Attention & $\mathrm{\Delta}$ & +0.9              & +0.3 & +0.3 & +0.6 & +4.0   & +1.4 & +1.0   & +3.0                & +1.9   & +1.9   & +2.0    & +3.8      \\
          & + $\mathrm{StrDA_{HDGE}}$     & \uline{97.2}              & \uline{95.2} & \textbf{\uline{96.5}} & \textbf{\uline{84.5}} & \uline{90.7} & \textbf{\uline{94.4}} & \uline{92.8} & \uline{68.6}              & \uline{42.7}   & \textbf{\uline{72.2}}   & \textbf{\uline{85.8}}  & \textbf{\uline{64.2}}      \\
          & $\mathrm{\Delta}$ & +1.0                & +1.5 & +0.7 & +2.6 & +4.6 & +3.4 & +1.8 & +6.1              & +3.7   & +3.2   & +3.0    & +7.6      \\
          & ABINet
  (baseline)   & 97.0                & 95.2 & 95.6 & 82.3 & 89.5 & 90.3 & 91.8 & 63.2              & 39.5   & 68.9   & 82.6  & 55.7      \\
          & + $\mathrm{ST}$           & 97.4              & 96.3 & \uline{96.4} & 83.9 & \textbf{\uline{91.0}}   & 92.0   & 92.8 & 68.7              & 42.3   & 71.2   & 84.7  & 61.4      \\
LM        & $\mathrm{\Delta}$ & +0.4              & +1.1 & +0.8 & +1.6 & +1.5 & +1.7 & +1.0   & +5.5              & +2.8   & +2.3   & +2.1  & +5.7      \\
          & + $\mathrm{StrDA_{HDGE}}$   & \textbf{\uline{97.8}}              & \textbf{\uline{96.9}} & 96.0   & \uline{84.4} & \textbf{\uline{91.0}}   & \textbf{\uline{94.4}} & \textbf{\uline{93.2}} & \textbf{\uline{69.7}}              & \textbf{\uline{44.2}}   & \uline{71.6}   & \uline{85.0}    & \uline{62.9}      \\
          & $\mathrm{\Delta}$ & +0.8              & +1.7 & +0.4 & +2.1 & +1.5 & +4.1 & +1.4 & +6.5              & +4.7   & +2.7   & +2.4  & +7.2      
\end{tblr}
\label{tab:main}
}
\end{table*}
\subsection{Stage 2: Progressive Self-Training}

At the end of Stage 1, we have $n$ subsets for Stage 2-Progressive Self-Training. As demonstrated in \cref{fig:Overview}, we will conduct self-training ($\mathrm{ST}$) sequentially on each set of sub-domain data $T_{i}$. The entire learning process is described in \cref{algo}.

\begin{algorithm}
  \caption{Progressive Self-Training $\mathrm{ST}$}
  \begin{algorithmic}[1]
    \Require Labeled images $(X, Y) \in S$ and sequence of unlabeled image subsets $T_{1}, T_{2}, T_{3},\ldots, T_{n} (T_i \in T)$
    \State Train STR model $M(\cdot, \theta_{0})$ with $(X, Y)$ using \cref{eq:STR}.
    \For{iteration i = 1, 2, $\ldots$, n}
        \State $T_i \rightarrow M(\cdot, \theta_{i-1}) \rightarrow V_i$ (pseudo-labels) and $m_i$ (average confidence-scores)
        \State Update $\theta_{i}$ with $(X,Y)$, $(T_{i},V_{i})$, $m_{i}$ using \cref{eq:total-loss}
    \EndFor
  \end{algorithmic}
  \label{algo}
\end{algorithm}

Given the input image $\bm{x}^L$ and the character sequence of the ground truth $\bm{y}^L ={y_1^L, \ldots, y_k^L}$, the STR model $M(\cdot;\theta)$ outputs a vector sequence $\textbf{p}^L = M(\bm{x}^L; \theta)$ $= {p_1^L, \ldots, p_k^L}$. Cross-entropy loss is employed to train the STR model:
\begin{equation}
    \begin{aligned}
    L_{\text{r}}(\bm{x}^L, \bm{y}^L) = \frac{1}{k} \sum_{i=1}^{k} \log p_{i}^L(y_{i}^L | \bm{x}^L)
    \label{eq:STR}
    \end{aligned}
\end{equation}
where $p_i^L(y_i^L)$ represents the predicted probability of the output being $y_i^L$ at time step $t$ and $k$ is the sequence length.

In each $\mathrm{ST}$ round, after obtaining labeled data and pseudo-labeled data, we proceed to train the STR model $M(\cdot;\theta)$ to minimize the objective function:
\begin{equation}
    \begin{aligned}
    L(\phi) & = \frac{1-m_i}{|S|}\sum_{\bm{x}^S \in S} L_{r}(\bm{x}^S; \bm{y}^S) \\ & + \frac{m_i}{|T_i|} \sum_{\bm{x}^{T_i} \in T_i}L_{r}(\bm{x}^{T_i}; \bm{y}^{T_i})
    \label{eq:total-loss}
    \end{aligned}    
\end{equation}
where $m_i$ is the mean (average) of confidence scores when generating pseudo-labels for the unlabeled image subset $T_i$. $m_i$ serves as an \textit{adaptive controller}.

\subsection{Additional Training Techniques}

\textbf{Label Sharpening.} We "sharpen" the soft labels to encourage the model to update its parameters. Consequently, during the training process in Stage 2, we utilize the model's predictions on unlabeled data as definitive pseudo-labels rather than relying on their probabilities.

\textbf{Regularization.} Regularization is a significant factor in self-training. Without regularization, the model is not incentivized to change during self-training \cite{kumar2020understanding}. Therefore, we also incorporate it into our model training process.

\textbf{Data Augmentation.} We apply multiple augmentation strategies on both geometry transformations and color jitter, which are borrowed from RandAugment \cite{cubuk2020randaugment}.

\section{Experiments}
\label{sec:exp}
\subsection{Datasets}
 
Our work focuses on addressing the domain gap problem between the source domain, which is \textbf{synthetic} data, and the target domain, which is \textbf{real} data in \textit{scene text recognition}.

Experiments are conducted according to the setup of \cite{baek2019wrong} to ensure a fair comparison. We used two types of data during the training process: synthetic data (with SynthText (ST) \cite{gupta2016synthetic} and MJSynth (MJ) \cite{jaderberg2014synthetic}) and real data without labels. Concretely, we collect public real-world datasets, including ArT \cite{chng2019icdar2019}, COCO-Text (COCO) \cite{veit2016coco}, LSVT \cite{sun2019icdar}, MLT19 \cite{nayef2019icdar2019}, OpenVINO \cite{krylov2021open}, RCTW17 \cite{shi2017icdar2017}, ReCTS \cite{zhang2019icdar}, Uber-Text (Uber) \cite{zhang2017uber}, TextOCR \cite{singh2021textocr}, and \textbf{discard their labels} to formulate the set of real data without labels. In addition, we exclude vertical text (height $>$ width) and images whose width is greater than 25 times the height. As a result, we have 16 million labeled synthetic data and 2 million unlabeled real data for training, denoted as \textbf{real unlabeled data (2M RU)}.

For evaluation, six standard benchmark datasets, including IIIT 5k-word (IIIT) \cite{mishra2012scene}, Street View Text (SVT) \cite{wang2011end},  ICDAR 2013 (IC13) \cite{karatzas2013icdar}, ICDAR 2015 (IC15) \cite{karatzas2015icdar}, SVT-Perspective (SVTP) \cite{phan2013recognizing}, and CUTE80 (CUTE) \cite{risnumawan2014robust} are used. Note that IC13 and IC15 have two versions of their respective test splits commonly used in the literature: 857 and 1,015 for IC13; 1,811 and 2,077 for IC15.

In order to achieve a comprehensive comparison, we expand our evaluation to encompass five larger and more challenging datasets: COCO-Text (COCO) \cite{veit2016coco}, Uber-Text (Uber) \cite{zhang2017uber}, ArT \cite{chng2019icdar2019}, ReCTS \cite{zhang2019icdar}, and Union14M \cite{jiang2023revisiting} (Artistic, Contextless, Curve, General).

\subsection{Evaluation Metrics}

Following standard conventions \cite{baek2019wrong}, we present word-level accuracy for each dataset. Furthermore, to provide a thorough evaluation of the models concerning their recognition performance on both \textit{regular} and \textit{irregular} text, as per \cite{baek2021if}, we introduce an average score denoted "Avg." This score represents the accuracy across the combined set of samples from all six benchmark datasets (IIIT\textsubscript{3000}, SVT\textsubscript{647}, IC13\textsubscript{1015}, IC15\textsubscript{2077}, SVTP\textsubscript{645}, and CUTE\textsubscript{288}).

\subsection{Implementation Details}

Three STR models, CRNN \cite{shi2016end}, TRBA \cite{baek2019wrong} and ABINet \cite{fang2021read}, are employed to assess the effectiveness of the proposed framework using their default configurations. We trained the \textit{baseline} STR models in a fully supervised manner using the synthetic dataset (MJ+ST). Our reproduced results from supervised training exceed those reported in the original papers \cite{shi2016end, baek2019wrong, fang2021read}. Besides the adopted augmentation techniques \cite{cubuk2020randaugment}, we trained the STR models for more iterations (300K).

For Stage 1 (Domain Stratifying), the Harmonic Domain Gap Estimator ($\mathrm{HDGE}$) utilizes a generative adversarial network. Details are in the supplementary materials.

For Stage 2 (Progressive Self-Training), we adopt the AdamW \cite{loshchilov2017decoupled} optimizer (weight\_decay 0.005). We also use the one-cycle learning rate scheduler \cite{smith2019super} with a maximum learning rate of 0.0005. Our training batch size is fixed at 128, and the total number of iterations is 50K.

To demonstrate the effectiveness of our StrDA approach compared to traditional unsupervised domain adaptation (vanilla self-training $\mathrm{ST}$) (\cref{sec:method}), we conducted all experiments with the same protocols. All experiments were performed on an NVIDIA RTX A5000 (24GB VRAM).

\begin{table*}[tb]
\caption{Comparison with other domain adaptation methods in STR task. Our method significantly enhances the performance of the STR model, surpassing other existing approaches. Additionally, it can be integrated with other methods to achieve even greater efficacy.}
\centering
\adjustbox{max width=\linewidth}
{
\begin{tblr}{
  row{5} = {c},
  row{6} = {c},
  row{7} = {c},
  row{8} = {c},
  row{10} = {c},
  row{11} = {c},
  row{12} = {c},
  cell{1}{2} = {r=2}{c},
  cell{1}{3} = {r=2}{c},
  cell{1}{4} = {r=2}{c},
  cell{1}{5} = {c=4}{c},
  cell{1}{9} = {c=4}{c},
  cell{2}{5} = {c},
  cell{2}{6} = {c},
  cell{2}{7} = {c=2}{c},
  cell{2}{9} = {c=2}{c},
  cell{2}{11} = {c},
  cell{2}{12} = {c},
  cell{3}{5} = {c},
  cell{3}{6} = {c},
  cell{3}{7} = {c},
  cell{3}{8} = {c},
  cell{3}{9} = {c},
  cell{3}{10} = {c},
  cell{3}{11} = {c},
  cell{3}{12} = {c},
  cell{4}{1} = {r=5}{},
  cell{4}{2} = {c},
  cell{4}{3} = {c},
  cell{4}{4} = {c},
  cell{4}{5} = {c},
  cell{4}{6} = {c},
  cell{4}{7} = {c},
  cell{4}{8} = {c},
  cell{4}{9} = {c},
  cell{4}{10} = {c},
  cell{4}{11} = {c},
  cell{4}{12} = {c},
  cell{9}{1} = {r=4}{},
  cell{9}{2} = {c},
  cell{9}{3} = {c},
  cell{9}{4} = {c},
  cell{9}{5} = {c},
  cell{9}{6} = {c},
  cell{9}{7} = {c},
  cell{9}{8} = {c},
  cell{9}{9} = {c},
  cell{9}{10} = {c},
  cell{9}{11} = {c},
  cell{9}{12} = {c},
  vline{8} = {2}{},
  vline{9} = {2-12}{},
  hline{1,13} = {-}{0.08em},
  hline{2} = {5-12}{},
  hline{4} = {-}{},
  hline{9} = {1-12}{},
  hline{9} = {2}{1-12}{},
  hline{10} = {2-12}{dashed},
}
                                                & Method                        & Labeled Dataset & Unlabeled Dataset   & Regular Text  &               &               &               & Irregular Text &               &                                                                              &               \\
                                                &                               &                 &                     & IIIT          & SVT           & IC13          &               & IC15           &               & SVTP                                                                         & CUTE          \\
                                                &                               &                 &                     & 3000          & 647           & 857           & 1015          & 1811           & 2077          & 645                                                                          & 288           \\
\begin{sideways}Published Results\end{sideways} & TRBA-FEDS \cite{patel2021feds}                     & MJ+ST           & Amazon\_book\_cover & 92.2          & 92.1          & 96.5          & 95.3          & 83.8           & 80.9          & 84.0                                                                         & 79.0          \\
                                                & TRBA-Seq-UPS \cite{patel2023seq}                  & MJ+ST           & 276K RU             & 92.7          & 88.6          & \_            & 92.2          & \_             & 76.9          & 78.0                                                                         & 84.4          \\
                                                & TRBA-cr \cite{zheng2022pushing}                       & MJ+ST           & 10.6M RU            & 96.5          & 96.3          & \textbf{98.3} & \_            & \textbf{89.3}  & \_            & \textbf{93.3} & 93.4          \\
                                                & ABINet-st \cite{fang2021read}                     & MJ+ST           & Uber-Text           & 96.8          & 94.9          & 97.3          & \_            & 87.4           & \_            & 90.1                                                                         & 93.4          \\
                                                & ABINet-est \cite{fang2021read}                    & MJ+ST           & Uber-Text           & 97.2          & 95.5          & 97.7          & \_            & 86.9           & \_            & 89.9                                                                         & 94.1          \\
\begin{sideways}Our Results\end{sideways}       & TRBA-cr (reproduce)           & MJ+ST           & 2M RU               & 97.3          & 95.1          & 97.2          & 96.2          & 88.1           & 84.0          & 90.5                                                                         & 93.8          \\
                                                & \textbf{TRBA-StrDA\textsubscript{HDGE}}       & MJ+ST           & 2M RU               & 97.2          & 95.2          & 97.4          & 96.5          & 88.4           & \textbf{84.5} & 90.7                                                                         & \textbf{94.4} \\
                                                & \textbf{TRBA-StrDA\textsubscript{HDGE} w/ cr} & MJ+ST           & 2M RU               & 97.3          & 96.1          & 97.6          & \textbf{96.7} & 88.7           & \textbf{84.5} & 90.9                                                                         & \textbf{94.4} \\
                                                & \textbf{ABINet-StrDA\textsubscript{HDGE}}     & MJ+ST           & 2M RU               & \textbf{97.8} & \textbf{96.9} & 97.0          & 96.0          & 88.6           & 84.4          & 91.0                                                                         & \textbf{94.4} 
\end{tblr}
\label{tab:SOTA}
}
\end{table*}

 \begin{table}[tb]
\centering
\caption{Effect of our proposed $\mathrm{HDGE}$. Compared to using $\mathrm{DD}$, StrDA with $\mathrm{HDGE}$ yields better results.}
\adjustbox{max width=\linewidth}
{
\begin{tblr}{
  column{even} = {c},
  column{3} = {c},
  column{5} = {c},
  column{7} = {c},
  cell{1}{1} = {b},
  cell{1}{8} = {r=2}{},
  hline{1,9} = {-}{0.08em},
  hline{3,5,7} = {-}{},
}
Method           & IIIT                  & SVT                   & IC13                  & IC15                  & SVTP                  & CUTE                  & Avg.                  \\
                 & 3000                  & 647                   & 1015                  & 2077                  & 645                   & 288                   &                       \\
CRNN-StrDA\textsubscript{DD}     & 93.3                  & 88.9                  & 92.5                  & 73.4                  & 76.7                  & 83.0                  & 85.7                  \\
CRNN-StrDA\textsubscript{HDGE}   & \uline{93.4}          & \uline{89.0}          & \uline{93.1}          & \uline{74.0}          & \uline{77.1}          & \uline{84.4}          & \uline{86.0}          \\
TRBA-StrDA\textsubscript{DD}     & \uline{97.6}          & 94.7                  & 96.1                  & 83.6                  & 89.9                  & 93.1                  & 92.5                  \\
TRBA-StrDA\textsubscript{HDGE}   & 97.2                  & \uline{95.2}          & \uline{\textbf{96.5}} & \uline{\textbf{84.5}} & \uline{90.7}          & \uline{\textbf{94.4}} & \uline{92.8}          \\
ABINet-StrDA\textsubscript{DD}   & \textbf{\uline{97.8}} & 96.8                  & \uline{96.2}          & 84.1                  & \textbf{\uline{91.0}} & 93.4                  & 93.0                  \\
ABINet-StrDA\textsubscript{HDGE} & \textbf{\uline{97.8}} & \textbf{\uline{96.9}} & 96.0                  & \uline{84.4}          & \textbf{\uline{91.0}} & \textbf{\uline{94.4}} & \textbf{\uline{93.2}} 
\end{tblr}
\label{tab:DD}
}
\end{table}

\subsection{Results and Analysis}

As illustrated in \cref{tab:main}, both domain adaptation methods ($\mathrm{ST}$ and $\mathrm{StrDA_{HDGE}}$) surpass the baseline models across eleven public benchmarks. Despite relying solely on additional unlabeled real data and self-training with pseudo-labels, the experiments remarkably enhanced the STR model's performance on both \textit{regular} and \textit{irregular} datasets. These remarkable results emphasize the importance of integrating real images into training STR models. 

Notably, the $\mathrm{StrDA_{HDGE}}$ method applied to all three baseline models of STR outperforms vanilla self-training ($\mathrm{ST}$). We observed that $\mathrm{ST}$ does not perform well without domain sequences, although it shows a slight improvement over the source-trained model (improved by 1.1\% for CRNN and 1\% for both TRBA and ABINet on Avg.). Our progressive self-training framework shows strong effectiveness by partitioning and organizing data according to the progressive increase in domain gap.

Specifically, CRNN, TRBA, and ABINet exhibit remarkable improvements on Avg. (\textbf{+1.6\%}, \textbf{+1.8\%} and \textbf{+1.4\%}) when applying $\mathrm{StrDA_{HDGE}}$. Furthermore, with large and challenging datasets such as Union14M, the $\mathrm{StrDA_{HDGE}}$ method demonstrates exceptional effectiveness (\textbf{+4.5\%}, \textbf{+7.6\%} and \textbf{+7.2\%}).  These results demonstrate the generalizability of our proposed methods, as $\mathrm{StrDA_{HDGE}}$ is effective across various STR models, including CTC-based, Attention-based, and LM-based models.

\subsection{Comparison with Other Methods}

In \cref{tab:SOTA}, we perform a comparative analysis of our proposed method with other unsupervised domain adaptation methods for scene text recognition. As we reimplemented the TRBA-cr method, the reproduced results were slightly different from those reported in the original paper \cite{zheng2022pushing}. This is because we used less data (2M compared to 10.6M) while keeping all other settings the same. 

TRBA-$\mathrm{StrDA_{HDGE}}$ achieved superior results in most datasets. Furthermore, when combining both methods, TRBA-$\mathrm{StrDA_{HDGE}}$ with \textit{cr}, the performance improved beyond what was achieved independently by either method. For other methods compared on the same backbone, $\mathrm{StrDA_{HDGE}}$ consistently demonstrated superior performance. The previous works are commendable. It is noteworthy that our framework can be conceptualized as a series of domain adaptation rounds. Consequently, integrating advanced techniques into stage 2 of our framework would be highly advantageous.

\subsection{Ablation Study}

In this section, we conduct a series of ablation experiments. TRBA and ABINet are used in all experiments due to their superior performance. The dataset used consists of 2M RU. Additional experiments are provided in the Supp.

\subsubsection{Ablation Study on Hyper-parameter $n$ in \cref{eq:domain}}

\begin{figure}[tb]
    \centering
    \includegraphics[width=\linewidth]{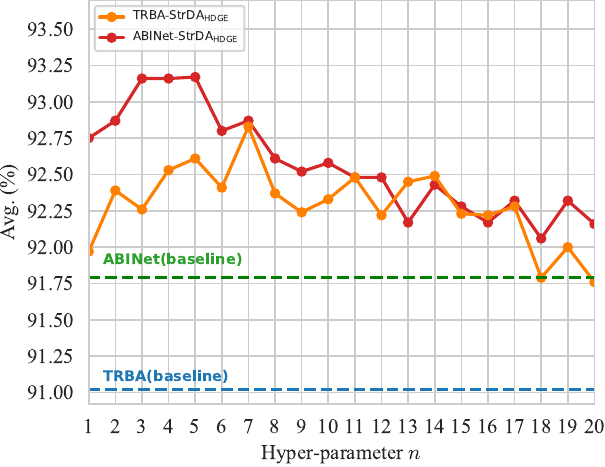}
    \caption{Ablation study on Hyper-parameter $n$ in \cref{eq:domain}.}
    \label{fig:groups}
\end{figure}

In this section, we proceed to compare the performance of $\mathrm{StrDA_{HDGE}}$ with different numbers of subsets (with the hyper-parameter $beta = 0.9$ and every subset has the same size). The case where $n = 1$ corresponds to vanilla self-training ($\mathrm{ST}$) as described in \cref{sec:method}.

According to \cref{fig:groups}, $\mathrm{StrDA_{HDGE}}$ proves to be more effective for both methods as the hyper-parameter $n$ increases, reaching \textbf{92.83\%} at $n=7$ for TRBA, and \textbf{93.17\%} at $n=5$ for ABINet. This phenomenon is understandable due to the reduction in domain disparity resulting from smaller data partitions, thereby facilitating the adaptability of the model. However, with larger $n$, the performance of the method tends to saturate and eventually decline, suggesting the need for a judicious choice of $n$. Future work should seek generalized approaches to determine $n$ and number of data samples in each subset.

\subsubsection{Ablation Study on Hyper-parameter $\beta$ in \cref{eq:hdge}}

\begin{figure}[tb]
    \centering
    \includegraphics[width=\linewidth]{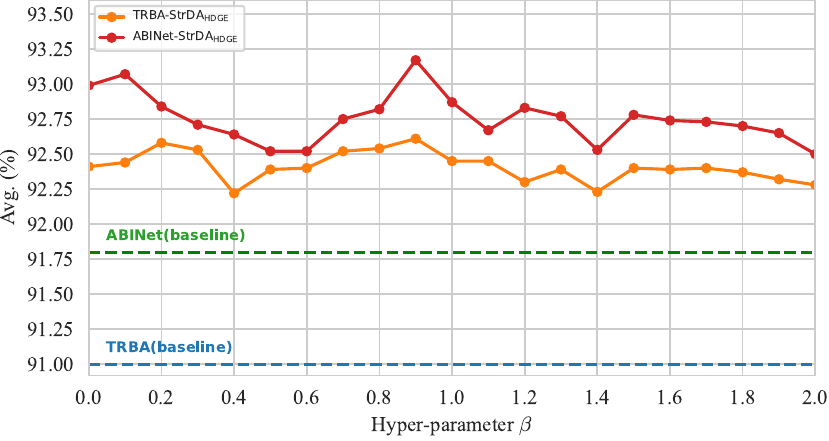}
    \caption{Ablation study on Hyper-parameter $\beta$ in \cref{eq:hdge}}
    \label{fig:beta}
\end{figure}

We conducted experiments (with the hyper-parameter $n=5$) to observe the influence of the hyper-parameter $\beta$ in \cref{eq:hdge} on the effectiveness of the StrDA\textsubscript{HDGE} method. We are adjusting the value of $\beta$ either higher or lower causing $\mathrm{HDGE}$ to exhibit less or more bias towards the source domain. When $\beta = 0$, $d_i = d_S$, $\mathrm{StrDA_{HDGE}}$ only uses information from the source domain. In this case, as shown in \cref{fig:beta}, StrDA demonstrates fairly good effectiveness (92.41\% for TRBA and 92.99\% for ABINet).

However, incorporating information from both the source and target directions leads to significantly higher performance (\textbf{92.61\%} and \textbf{93.17\%} for $\beta = 0.9$). This supports our suggestion in \cref{eq:hdge}. It also reinforces our claim that stratifying domain gaps using information from both source and target domains contributes to overall effectiveness.

\subsubsection{Alternative Domain Gap Estimators}

We conducted additional experiments with an alternative gap estimator, called Domain Classifier ($\mathrm{DD}$), to assess the domain gap $d_i$ (\cref{method:stage1}).
$\mathrm{DD}$ employs a binary classifier $f(\bm{x_i};\phi)$ with a feature extractor from the baseline model combined with a fully connected layer at the final layer. $\mathrm{DD}$ is trained using raw images from $S$ (assigned as class 0) and $T$ (assigned as class 1). Next, we assign $d_i = f(\bm{x_i};\phi)$. By learning distinctive features from the two domains, the discriminator identifies whether a data point is closer to the source or target domain, corresponding to a smaller or larger distance from the source.

As shown in \cref{tab:DD}, $\mathrm{StrDA_{HDGE}}$ yields better results compared to $\mathrm{StrDA_{DD}}$. We note that $\mathrm{DD}$ treats data points situated in the intersection and those outside both distributions similarly, with the same $d_i$. This leads to poor discrimination in the self-learning process.

\section{Conclusion}
\label{sec:conclu}
\begin{figure}[tb]
    \centering
    \includegraphics[width=\linewidth]{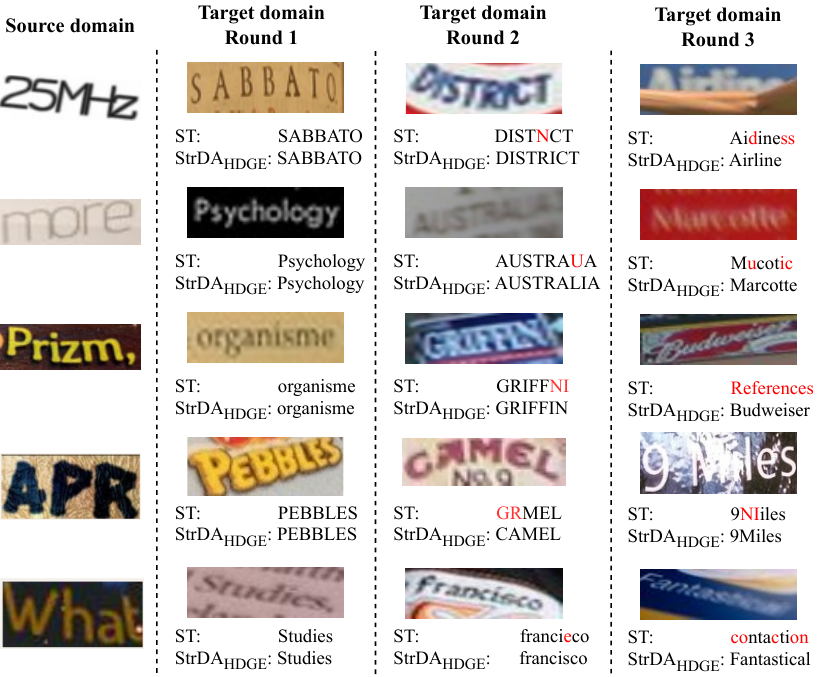}
    \caption{The StrDA partitions the data from the target domain into three distinct subsets, with the disparity across domains gradually rising, as shown in the image. The next two lines depict the pseudo-labels employed in the self-training process of $\mathrm{ST}$ and $\mathrm{StrDA_{HDGE}}$, respectively. The pseudo-labels generated by $\mathrm{ST}$ are prone to noise as the extent of the domain gap escalates. On the other hand, $\mathrm{StrDA_{HDGE}}$, produces pseudo-labels with higher accuracy. The STR model used for the example is TRBA.}
    \label{fig:results}
\end{figure}

In this paper, we propose the Stratified Domain Adaptation (StrDA) approach, a progressive self-training framework for scene text recognition. By leveraging the gradual escalation of the domain gap with the Harmonic Domain Gap Estimator ($\mathrm{HDGE}$), we propose partitioning the target domain into a sequence of ordered subsets to progressively reduce the domain gap between each and the source domain. Progressive self-training is then applied sequentially to these subsets. Extensive experiments on STR benchmarks demonstrate that our approach enables the baseline STR models to progressively adapt to the target domain. This approach significantly improves the performance of the baseline model without using any human-annotated data and shows its superior effectiveness compared to existing UDA methods for the scene text recognition task.

\section*{Acknowledgements}
This research is funded by University of Information Technology - VNUHCM under grant number D4-2024-03.

{\small
\bibliographystyle{ieee_fullname}
\bibliography{egbib}
}

\clearpage
\setcounter{section}{0}

\begin{figure}[h]
    \centering
    \includegraphics[width=\columnwidth]{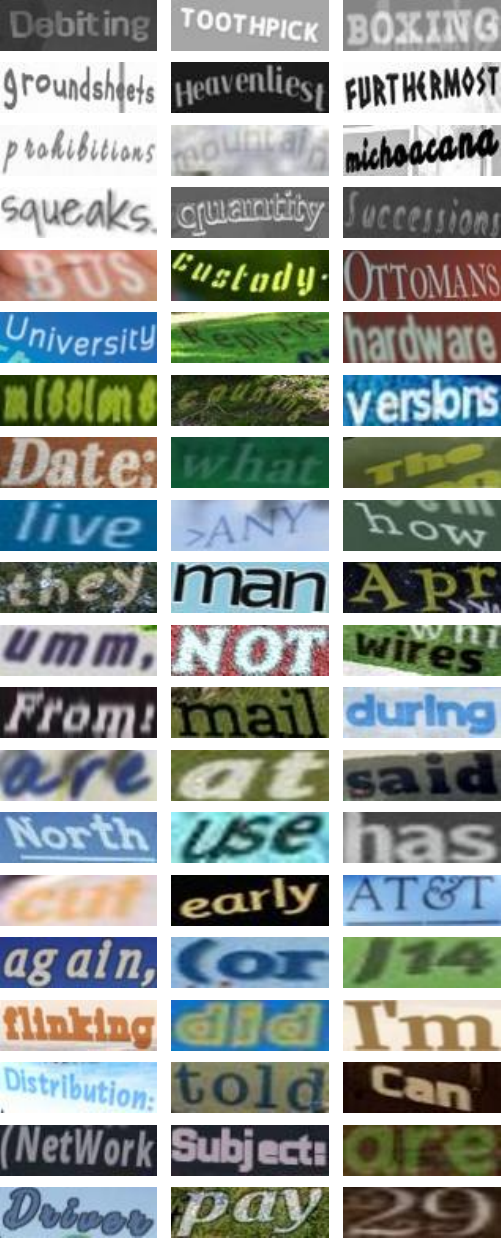}
    \caption{Examples of synthetic data. The samples are extracted from the MJ and ST datasets.}
    \label{fig:source}
\end{figure}

\section{Dataset Descriptions}

Our approach leverages \textit{labeled synthetic data} and \textit{unlabeled real data}, as shown in \cref{tab:dataset}. We \textbf{discard the labels} of real datasets to align with the experiments. The "Train." data we report is slightly different from \cite{baek2021if, bautista2022scene} because we use raw images (with discarded labels).

\begin{table*}[h]
\centering
\caption{Summary of dataset usage. Numbers indicate how many samples were used from each dataset. "t" refers to splits that were repurposed as training data. "*" note that we use the Union14M-Benchmark, which comprises: Artistic, Contextless, Curve, and General.}
\adjustbox{max width=\textwidth}
{
\begin{tabular}{lllrrr} 
\toprule
\multirow{2}{*}{Dataset} & \multirow{2}{*}{Conf.}                   & \multirow{2}{*}{Year} & \multicolumn{3}{c}{\# of word boxes}                                                                  \\ 
\cmidrule(l){4-6}
                         &                                       &                       & Train.    & Val.                          & Eval.                                                     \\ 
\hline
\multicolumn{6}{l}{\textbf{\textbf{Synthetic datasets}}}                                                                                                                                            \\
~ MJ \cite{jaderberg2014synthetic}                    & \textcolor[rgb]{0.122,0.133,0.161}{NIPSW} & 2014                  & 7,224,586 & 802,731\textsuperscript{t}    & \textcolor[rgb]{0.2,0.2,0.2}{891,924\textsuperscript{t}}  \\
~ ST \cite{gupta2016synthetic}                     & CVPR                                     & 2016                  & 6,975,301 & -                             & -                                                         \\ 
\hline\hline
\multicolumn{6}{l}{\textbf{\textbf{Real datasets}}}                                                                                                                                \\
~ IIIT5k \cite{mishra2012scene}                & BMVC                                     & 2012                  & 2,000     & -                             & \textcolor[rgb]{0.2,0.2,0.2}{3,000}                       \\
~ SVT \cite{wang2011end}                   & ICCV                                     & 2011                  & 257       & -                             & \textcolor[rgb]{0.2,0.2,0.2}{647}                         \\
~ IC13 \cite{karatzas2013icdar}                  & ICDAR                                    & 2013                  & 848       & -                             & \textcolor[rgb]{0.2,0.2,0.2}{1,015}                 \\
~ IC15 \cite{karatzas2015icdar}                  & ICDAR                                    & 2015                  & 4,468     & -                             & \textcolor[rgb]{0.2,0.2,0.2}{2,077}               \\
~ SVTP \cite{phan2013recognizing}                  & ICCV                                     & 2013                  & -         & -                             & \textcolor[rgb]{0.2,0.2,0.2}{645}                         \\
~ CUTE \cite{risnumawan2014robust}                  & ESWA                                     & 2014                  & -         & -                             & 288                                                       \\
~ COCO \cite{veit2016coco}                  & arXiv                                    & 2016                  & 59,820    & ~ ~13,415  & \textcolor[rgb]{0.2,0.2,0.2}{9,825}                       \\
~ Uber \cite{zhang2017uber}                  & CVPRW                                    & 2017                  & 91,978    & 36,136     & 80,418                                                    \\
~ ArT \cite{chng2019icdar2019}                   & ICDAR                                    & 2019                  & 32,349    & -                             & \textcolor[rgb]{0.2,0.2,0.2}{35,149}                      \\
~ ReCTS \cite{zhang2019icdar}                 & ICDAR                                    & 2019                  & 25,328    & -                             & ~ ~2,592                                                  \\
~ LSVT \cite{sun2019icdar}                  & ICDAR                                    & 2019                  & 43,244    & -                             & -                                                         \\
~ MLT19 \cite{nayef2019icdar2019}                 & ICDAR                                    & 2019                  & 56,937    & -                             & -                                                         \\
~ RCTW17 \cite{shi2017icdar2017}                & ICDAR                                    & 2017                  & 10,509    & -                             & -                                                         \\
~ TextOCR \cite{singh2021textocr}               & ECCV                                     & 2020                  & 714,770   & ~ ~107,722 & -                                                         \\
~ OpenVINO \cite{krylov2021open}               & ACML                                     & 2021                  & 1,914,425   & ~ ~158,819   & -                                                         \\
~ Union14M-Benchmark* \cite{jiang2023revisiting}               & ICCV                                     & 2023                  & -   & -   & 403,379                                                         \\

\bottomrule
\end{tabular}
}
\label{tab:dataset}
\end{table*}

We present some data from the source domain (synthetic) in \cref{fig:source}. Compared to the target domain in \cref{fig:example}, a significant domain gap appears between the two domains, affecting the performance of the STR models.

\section{Harmonic Domain Gap Estimator (HDGE) details}

To create a pair of discriminators $D_S$ and $D_T$ with the ability to assess out-of-distribution (OOD) levels effectively, we used a learning strategy inspired by \cite{zhu2017unpaired, zhan2019ga}. Our discriminators ($D_S$ and $D_T$) are described in \cref{tab:discriminator}.

\section{Domain Discriminator (DD) details}

\subsection{Training detail (stage 1)}

Domain Discriminator ($\mathrm{DD}$) employs a binary classifier $f(\bm{x};\phi)$ with a feature extractor from the baseline model combined with a fully connected layer at the last layer. $\mathrm{DD}$ is trained with raw images from $S$ (assigned as class 0) and $T$ (assigned as class 1).

We use focal loss \cite{lin2017focal} to optimize the learnable parameter to improve DD's accuracy in classifying challenging cases and addressing data imbalance issues (\eg class 0 with 16 million samples and class 1 with 2 million data samples):
\begin{equation}
   \begin{aligned}
   L(\phi) &= -\frac{1}{|S|}\sum_{\bm{x}^S \in S} (\sigma(f(\bm{x}^{S}; \phi)))^{\gamma} \log(1-\sigma(f(\bm{x}^S; \phi))) \\&-\frac{1}{|T|} \sum_{\bm{x}^{T} \in T}(1-\sigma(f(\bm{x}^{T} ; \phi)))^\gamma \log (\sigma(f(\bm{x}^{T}, \phi)))
   \end{aligned}
\end{equation}
where $\sigma$ is the sigmoid function. Then, we assign $d_{i}=\sigma(f(\bm{x}_{i};\phi)), d_{i} \in (0,1)$ to a data point $\bm{x}_{i}^T$. The focusing hyper-parameter $\gamma$ smoothly adjusts the rate at which easy examples are down-weighted.

\begin{table}[tb]
\centering
\caption{Discriminator ($D_S$ and $D_T$) architecture configuration for the Harmonic Domain Gap Estimator. Here, c, k, s, and p stand for no. of channels, filter size, stride, and padding, respectively.}
\adjustbox{max width=\columnwidth}
{
\begin{tabular}{|c|c|c|}
\hline
\textbf{Layers} & \textbf{Configurations}          & \textbf{Output} \\ \hline
Input           & image                            & 100x32x3        \\ \hline
Conv1           & c: 64,   k: 4x4,   s: 2,  p: 1   & 50x16x64        \\ \hline
Activation      & Leaky ReLU (0.2)                 & 50x16x64        \\ \hline
Conv2           & c: 128,   k: 4x4,   s: 2,   p: 1 & 25x8x128        \\ \hline
Activation      & Leaky ReLU (0.2)                 & 25x8x128        \\ \hline
Conv3           & c: 256,   k: 4x4,   s: 2,   p: 1 & 12x4x256        \\ \hline
Activation      & Leaky ReLU (0.2)                 & 12x4x256        \\ \hline
Conv4           & c: 512,   k: 4x4,  s: 1,   p: 1  & 11x3x512        \\ \hline
Activation      & Leaky ReLU (0.2)                 & 11x3x512        \\ \hline
Conv5           & c: 1,   k: 4x4,   s: 1,   p: 1   & 10x2x1          \\ \hline
\end{tabular}
}
\label{tab:discriminator}
\end{table}

\subsection{Ablation Study on DD (stage 2)}

\begin{figure}[tb]
    \centering
    \includegraphics[width=\columnwidth]{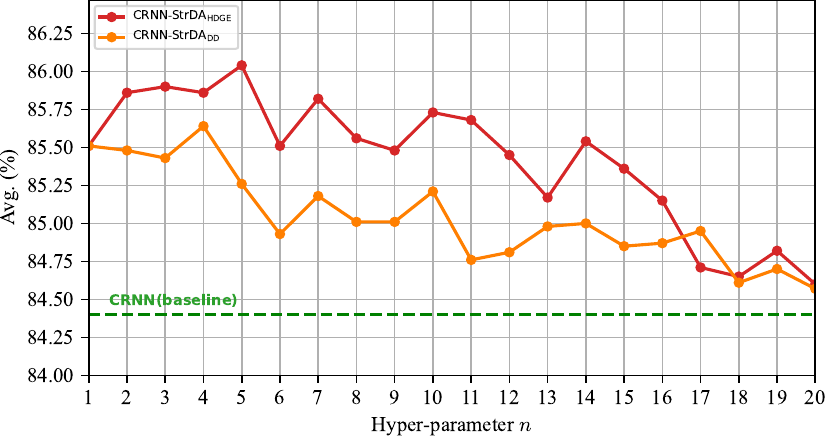}
    \caption{Ablation study on the hyper-parameter $n$ for CRNN-StrDA\textsubscript{HDGE} and CRNN-StrDA\textsubscript{DD}.}
    \label{fig:crnn_dd}
\end{figure}

\begin{figure}[tb]
    \centering
    \includegraphics[width=\columnwidth]{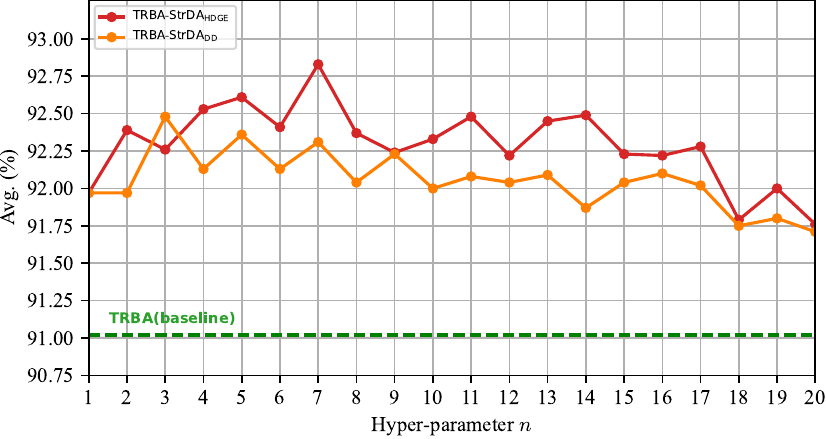}
    \caption{Ablation study on the hyper-parameter $n$ for TRBA-StrDA\textsubscript{HDGE} and TRBA-StrDA\textsubscript{DD}.}
    \label{fig:trba_dd}
\end{figure}

\begin{figure}[tb]
    \centering
    \includegraphics[width=\columnwidth]{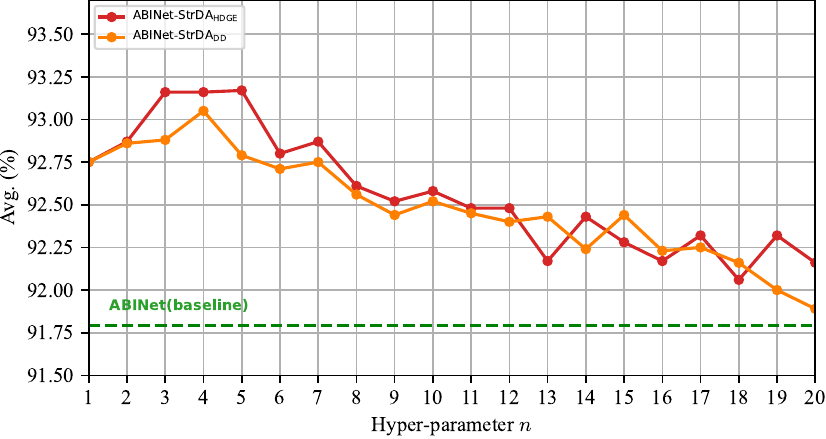}
    \caption{Ablation study on the hyper-parameter $n$ for ABINet-StrDA\textsubscript{HDGE} and ABINet-StrDA\textsubscript{DD}.}
    \label{fig:abinet_dd}
\end{figure}

We experimented with the method $\mathrm{StrDA_{DD}}$ using various settings for the hyper-parameter $n$. As shown in \cref{fig:crnn_dd}, \cref{fig:trba_dd}, and \cref{fig:abinet_dd}, in most cases, $\mathrm{StrDA_{HDGE}}$ demonstrates superior performance compared to $\mathrm{StrDA_{DD}}$. Moreover, as hyper-parameter $n$ is too high, the effectiveness of StrDA decreases. Therefore, a reasonable choice of $n$ is crucial. Future work could explore optimal methods for selecting $n$.

\section {Qualitative Results}

\begin{figure}[tb]
    \centering
    \includegraphics[width=\columnwidth]{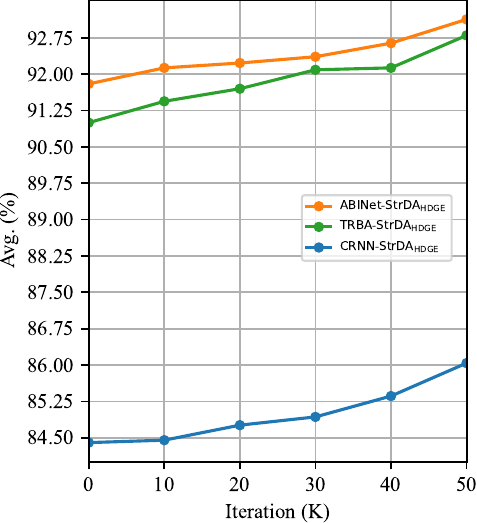}
    \caption{The stability of the STR models throughout the progressive self-training process. It can be observed that the accuracy of the TRBA model steadily increases across rounds.}
    \label{fig:stable}
\end{figure}

In \cref{fig:stable}, we visualize the performance of the STR models during the progressive self-training process. $\mathrm{StrDA_{HDGE}}$ shows improved performance, and the stability of the STR models is reinforced throughout each round of progressive self-training.

\begin{figure*}[tb]
    \centering
    \includegraphics[width=0.7\textwidth]{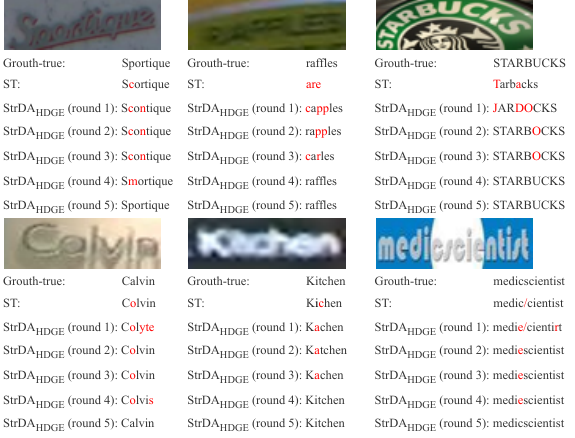}
    \caption{Predictions of TRBA-StrDA\textsubscript{HDGE} model on some cases from the benchmark dataset after each round of self-training. It can be seen that the model gradually improves its accuracy compared to the previous round. Misclassified characters are highlighted in \textcolor{red}{red}.}
    \label{fig:acc}
\end{figure*}

In \cref{fig:acc}, we observe the predictions of the TRBA-StrDA\textsubscript{HDGE} model in some cases from benchmark datasets. After progressive self-training, the TRBA model gradually improves its accuracy compared to the previous round.

\begin{figure*}[h]
    \centering
    \includegraphics[width=\textwidth]{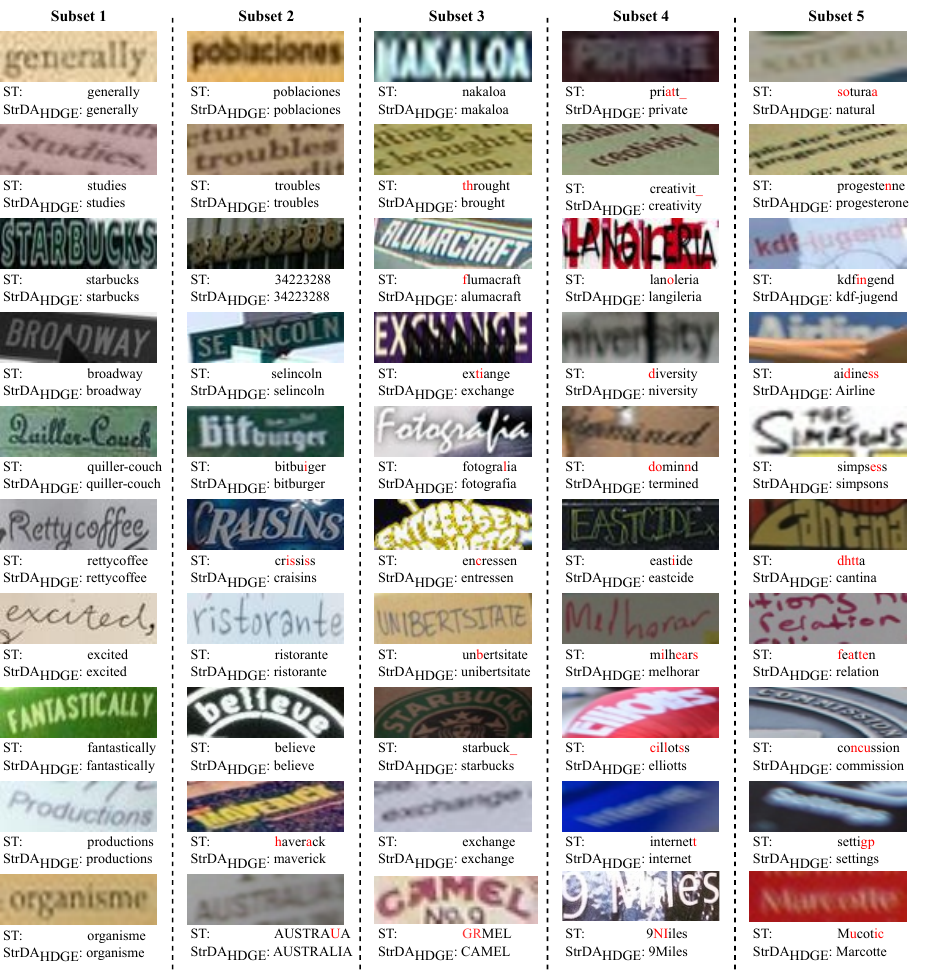}
    \caption{The Stratified Domain Adaptation ($\mathrm{StrDA_{HDGE}}$) approach partitions the data from the target domain into five distinct subsets, with the disparity across domains gradually increasing, as shown in the image. The difficulty of challenging cases (curved or perspective texts, occluded texts, texts in low-resolution images, and texts written in difficult fonts) increases progressively across these subsets. The subsets are then subjected to self-training in sequential rounds. We observe the pseudo-labels generated by the TRBA model for each subset at the beginning of the self-training process. In the case of vanilla self-training ($\mathrm{ST}$), all cases are predicted simultaneously by the source-trained (baseline) model. In $\mathrm{StrDA_{HDGE}}$, the model predicts pseudo-labels for the target domain in round $m$ using the TRBA model after self-training in round $m-1$. The pseudo-labels generated by $\mathrm{ST}$ are prone to noise (\textcolor{red}{red} characters) as the extent of the domain gap escalates. On the other hand, $\mathrm{StrDA_{HDGE}}$ produces pseudo-labels with higher quality. This contributes to making the progressive self-training process much more effective. The STR model used for the example is TRBA.}
    \label{fig:example}
\end{figure*}

To visually observe how StrDA operates, we sampled some cases from each subset after partitioning. As illustrated in \cref{fig:example}, the difficulty of challenging cases increases gradually through each round. Therefore, when applying progressive self-training to the TRBA model, the recognizer can adapt progressively across each subset from the source to the target domain. $\mathrm{StrDA_{HDGE}}$ also demonstrates superior performance in generating high-quality pseudo-labels compared to vanilla self-training $\mathrm{ST}$.

\end{document}